\documentclass{vgtc}

\ifpdf
  \pdfoutput=1\relax
  \usepackage{graphicx}
  \DeclareGraphicsExtensions{.pdf,.png,.jpg,.jpeg}
\else
  \usepackage{graphicx}
  \DeclareGraphicsExtensions{.eps}
\fi

\graphicspath{{figures/}{pictures/}{images/}{./}}

\usepackage{microtype}
\PassOptionsToPackage{warn}{textcomp}
\usepackage{textcomp}
\usepackage{mathptmx}
\usepackage{times}

\usepackage{cite}
\usepackage{tabu}
\usepackage{booktabs}
\usepackage{amsmath}
\usepackage{amssymb}
\usepackage{wrapfig}
\usepackage{xspace}
\usepackage{hyperref}
\usepackage{enumitem}
\usepackage{graphicx}
\usepackage{xcolor,soul}

\onlineid{0}
\vgtccategory{Research}
\vgtcinsertpkg

\title{Bluff: Interactively Deciphering Adversarial Attacks on \\Deep Neural Networks}

\author{
    Nilaksh Das
        \thanks{
            Authors contributed equally.
        }$\,\,\,\,^\dagger$, 
    Haekyu Park$^{\ast\dagger}$,
    Zijie J. Wang
        \thanks{
            Georgia Institute of Technology. 
            \{nilakshdas, haekyu, jayw, fredhohman, rfirstman6, polo\}@gatech.edu
        }$\,\,\,$,
    Fred Hohman$^\dagger$,\\
    Robert Firstman$^\dagger$,
    Emily Rogers
        \thanks{
            Georgia Tech Research Institute.
            Emily.Rogers@gtri.gatech.edu 
        }$\,\,\,$,
    Duen Horng (Polo) Chau$^\dagger$
}

\newcommand{\modeltitle}[0]{Bluff}
\newcommand{\model}[0]{\textsc{\modeltitle}}
\newcommand{\sidebar}[0]{Control Sidebar\xspace}
\newcommand{\graphview}[0]{Graph Summary View\xspace}
\newcommand{\detailview}[0]{Detail View\xspace}

\newcommand{\inception}{\textsc{InceptionV1}\xspace}

\definecolor{green}{RGB}{50,149,54}
\definecolor{orange}{RGB}{255,143,40}
\definecolor{red}{RGB}{198,50,42}
\definecolor{agreen}{RGB}{74, 198, 148}
\definecolor{purple}{RGB}{158, 62, 177}
\definecolor{darkpurple}{RGB}{170, 70, 210}
\definecolor{aqua}{RGB}{87, 180, 181}
\definecolor{lightblue}{RGB}{67, 130, 181}

\newlength{\myMheight}
\newcommand{\blue}[1]{\textcolor{lightblue}{#1}}
\newcommand{\red}[1]{\textcolor{red}{#1}}
\newcommand{\orange}[1]{\textcolor{orange}{#1}}

\newcommand{\green}[1]{\textcolor{green}{#1}}

\teaser{
  \centering
  \includegraphics[width=0.9\linewidth]{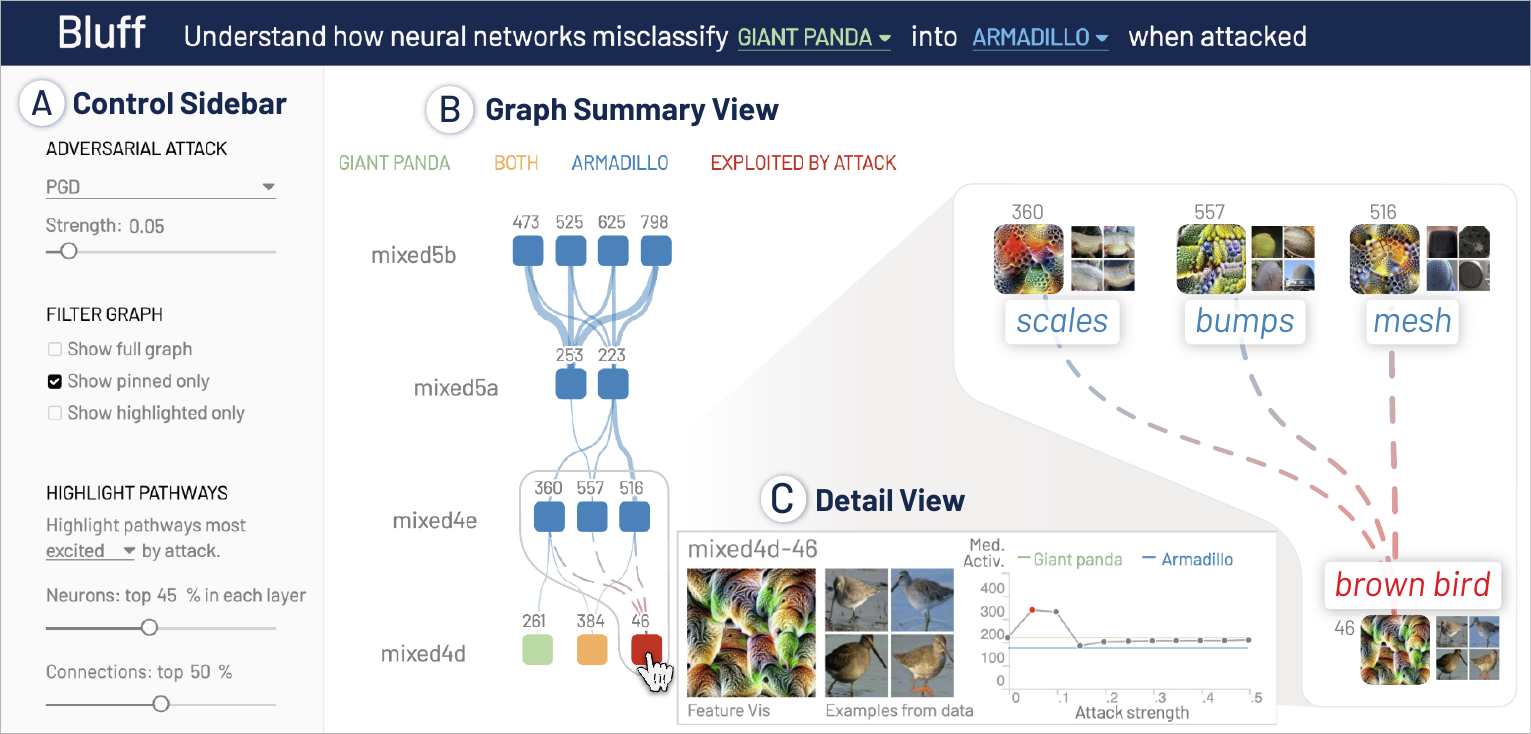}
  \vspace{-0.6em}
  \caption{
    With \model{}, users interactively visualize 
    how adversarial attacks penetrate a deep neural network to induce incorrect outcomes.
    Here, a user inspects why \inception{} misclassifies adversarial \green{\textbf{giant panda}} images, crafted by the \textit{Projected Gradient Descent} (PGD) attack, as \blue{\textbf{armadillo}}.
    PGD successfully perturbed pixels to induce the ``\red{\textit{brown bird}}'' feature, an appearance more likely shared by an armadillo (small, roundish, brown body) than a panda, 
    activating more features 
    that contribute to the armadillo (mis)classification (e.g., ``\blue{\textit{scales}},'' ``\blue{\textit{bumps}},'' ``\blue{\textit{mesh}}''). 
    The \textit{adversarial} pathways, formed by these neurons and their connections, overwhelm the benign panda pathways and lead to the ultimate misclassification.
    \textbf{(A)~\sidebar{}} allows users to specify what data is to be included
    and highlighted.
    \textbf{(B)~\graphview{}} visualizes pathways 
    most activated or changed by an attack as a network graph
    of neurons (each labeled by the channel ID in its layer) and their connections.
    When hovering over a neuron, 
    \textbf{(C)~Detail View} displays its feature visualization, representative dataset examples, and activation patterns over attack strengths. 
  }
  \label{fig:teaser}
}

\abstract{Deep neural networks (DNNs) are now commonly used in many domains.
However, they are vulnerable to \textit{adversarial attacks}: carefully-crafted perturbations on data inputs that can fool a model into making incorrect predictions.
Despite significant research on developing DNN attack and defense techniques, 
people still lack an understanding of how such attacks penetrate a model's internals.
We present \model{}, an interactive system for visualizing, characterizing, and deciphering adversarial attacks on vision-based neural networks. 
\model{} allows people to flexibly visualize and compare the activation pathways for benign and attacked images, revealing  mechanisms that adversarial attacks employ to inflict harm on a model.
\model{} is open-sourced and runs in modern web browsers.
} 

\CCScatlist{
  \CCScatTwelve{Human-centered computing}{Visual Analytics}{}
}

\begin{document}

\maketitle

\section{Introduction}

Deep neural networks (DNNs) are a major driving force behind many recent technological breakthroughs 
\cite{esteva2019guide,grigorescu2019survey,guo2019survey,nassif2019speech,heaton2016deep,zhang2019deep},
but they are highly vulnerable to \textit{adversarial attacks}. 
Small, human-imperceptible noise injected into inputs can easily fool DNNs into making wrong predictions
\cite{chen2018shapeshifter,Goodfellow2014ExplainingAH,kurakin2016adversarial,qin2019imperceptible},
raising alarms for
safety-critical applications, such as autonomous driving and data-driven healthcare.
Thus, it is essential to understand how attacks harm DNN models \cite{ross2018improving,tao2018attacks}. 
But interpreting and ultimately defending against adversarial attacks remain fundamental research challenges.
DNNs are often considered ``unintelligible'' due to their complex architectures and huge number of parameters.
It is difficult to pinpoint the parts of the model 
exploited by an attack,
let alone to understand how such exploitation leads to incorrect outcomes \cite{liu2018analyzing}.
Also, there is a lack of research in understanding how 
an attack's ``strength'' may correlate with neurons' activation patterns
\cite{madry2018adversarial}.
For example, it is not yet known if
a stronger attack exploits the same neurons as a weaker attack does, or if these sets are completely different.

\begin{figure}[t]
    \centering
    \includegraphics[width=0.9\linewidth]{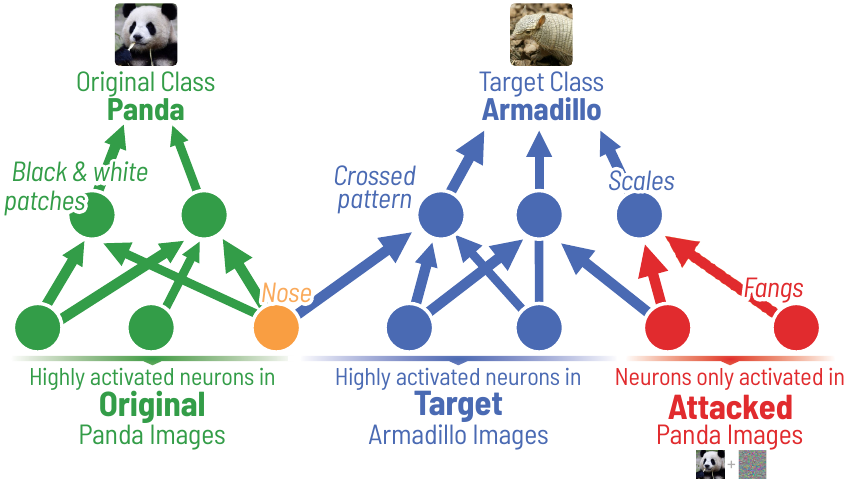}
    \vspace{-1em}
    \caption{
        Adversarial attacks confuse DNNs to make incorrect predictions (e.g., misclassify benign \textit{panda} as \textit{armadillo}).
        \model{} helps discover where such attacks occur 
        and what features are used.
    }
    \label{fig:intro-compare}
    \vspace{-1em}
\end{figure}

To address the above challenges, we develop \model{} (\autoref{fig:teaser}), an interactive visualization tool for discovering and interpreting how adversarial attacks mislead DNNs into making incorrect decisions.
Our main idea is to visualize \textbf{activation pathways} within a DNN traversed by the signals of benign and adversarial inputs.
An \textit{activation pathway} consists of neurons (also called \textit{channels} or \textit{features}) that are highly activated or changed by the input, and the connections among the neurons.
\model{} finds and visualizes where a model is exploited by an attack, 
and what impact the exploitation has on the final prediction, across multiple attack strengths.
We contribute: 

\begin{itemize}[topsep=0.5mm, itemsep=0mm, parsep=0.5mm, leftmargin=3mm]
\item \textbf{\model{}, an interactive system for summarizing and interpreting} how adversarial perturbations penetrate DNNs to induce incorrect outcomes in \inception{} \cite{szegedy2015going}, a large-scale prevalent image-classification model, over images from ImageNet ILSVRC 2012 \cite{ILSVRC15}.
To support reproducible  research and broaden its access, 
we have open-sourced \model{} at
\href{https://poloclub.github.io/bluff}{\texttt{\blue{https://poloclub.github.io/bluff}}}.

\item \textbf{Visual characterization of activation pathway dynamics.}
Adversarial perturbations manipulate activation pathways typically used for benign inputs to induce incorrect predictions.
For example, an attack can \textit{inhibit} neurons detecting important features for the benign class and
\textit{excite} those that exacerbate misclassification.
\model{} visualizes and highlights activation pathways exploited by an attack (\autoref{fig:intro-compare}) and shows how they mutate and propagate through a network.

\item \textbf{Interactive comparison of attack escalation.}
\model{} enables interactive comparison of activation pathways under increasing attack strengths, providing a new way for understanding the essence of an attack (e.g., common trends of an attack across all strengths) and its multi-faceted characteristics (e.g., various strategies that different strengths may employ).

\item \textbf{Discovery usage scenarios.}
We describe how \textsc{Bluff} can help discover 
surprising insights into the vulnerability of DNNs, such as 
how unusual activation pathways may be exploited by attacks.

\end{itemize}

\section{Related Work}

\textbf{Adversarial Attacks on DNNs.}
Adversarial attacks aim to confuse a DNN model into making incorrect predictions by adding carefully crafted perturbations to the input ~\cite{Goodfellow2014ExplainingAH,moosavi2016deepfool,carlini2017towards}.
We focus on \textit{targeted} adversarial attacks, 
where a model is misled to make a prediction of the attacker's choosing,
which can pose severe threats to practical deep learning applications~\cite{chen2018shapeshifter}.
Given a benign input instance $x$, a targeted adversarial attack 
aims to find a small \textit{perturbation} $\delta$ 
that changes the prediction of model $\mathcal{M}$ 
to a target class $t$ different from the true class $y$, i.e.,
$\mathcal{M}(x+\delta) = t$, where $t\neq y, ||\delta||\le\epsilon$. 
We call $\epsilon$ the \textit{attack strength}.
Projected Gradient Descent (PGD)~\cite{madry2018adversarial} 
is one of the \textit{strongest} first-order targeted attacks 
~\cite{wang2019convergence}.
Hence, we examine the $l_2$ norm of PGD
with varying the strength $\epsilon$ from 0.0 (no attack) to 0.5 (strong attack).

\smallskip
\noindent
\textbf{Neural Network Interpretability.}
Deep neural networks have often been described as ``black boxes'' 
due to their complex internal structures. 
An approach to understand how neural networks work internally is
to study neurons' activation patterns.
To interpret what concept a neuron is detecting,
\textit{feature visualization} 
\cite{olah2017feature, mordvintsev2015inceptionism, carter2019activation, erhan2009visualizing} 
creates visualization that maximizes such neuron.
TCAV \cite{kim2017interpretability}, Network Dissection \cite{bau2017network}, 
and Net2Vec \cite{fong2018net2vec} propose to quantify interpretability by measuring alignment 
between the neuron activations and concept features.
Circuits \cite{olah2020zoom} and Summit \cite{hohman2019summit} visually explain 
how higher-level concepts can be constructed
by neural connections.
Activation Atlas~\cite{carter2019activation}
visualizes neuron activations per layer and analyzes how models can be exploited when predicting on manipulated inputs.
On top of using the neurons' activations, 
we visualize important connections among the neurons 
contributing to such misclassifications, 
and how these connections react to attacks.

\smallskip
\noindent
\textbf{Interpretability for Adversarial Attacks.}
While research on machine learning security has attracted great attention 
\cite{chen2018shapeshifter, das2018shield, madry2018adversarial, Goodfellow2014ExplainingAH, tramer2017ensemble, norton2017adversarial},
research for interpreting adversarial attacks on DNNs is nascent.
AEVis \cite{liu2018analyzing, cao2020analyzing} 
proposes to extract critical neurons and their connections 
for benign and adversarial inputs, 
and demonstrates the method on small sets of images.
However, it is unclear how it may scale to larger datasets that 
\model{} operates on 
(e.g., 900+ adversarial images for a single ImageNet class \cite{ILSVRC15}).
\model{} also provides new techniques for comparing activation pathways, 
enabling novel analysis (e.g., based on neuron inhibition and excitation) 
and discoveries (e.g., how different attacks may have different strategies).

\section{\model{}: Deciphering Adversarial Attacks}

\subsection{Design Goals}
\label{sec:challenges}
Through a literature survey, we have identified the following four design goals (\textbf{G1}-\textbf{G4}) that guide \model{}'s development.

\begin{enumerate}[label=\textbf{G\arabic*},itemsep=0mm, topsep=1mm,parsep=1mm, leftmargin=6mm]

    \item \label{challenge:entangled-pathways}
        \textbf{Untangling activation pathways.}
        \textit{Benign} activation pathways can significantly overlap with \textit{adversarial} pathways, as some neurons are ``\textit{polysemantic},'' detecting multiple concepts at the same time \cite{olah2020zoom, bau2017network}. 
        We aim to identify neurons that respond differently between benign and attacked inputs,
        to help discover where and how a model is exploited by an attack to induce incorrect predictions.
    \item \label{challenge:comparison}
        \textbf{Interpreting multiple activation pathways.}
        Understanding the effects of adversarial attacks is core to developing robust defenses 
        \cite{Goodfellow2014ExplainingAH, tramer2017ensemble, madry2018adversarial}.
        We aim to visualize high-level overviews of \textit{benign} and \textit{adversarial} activation pathways, and support drilling-down into subpaths,
        to help shed light on how specific groups of neurons are exploited to inflict harm on a model.
    \item \label{challenge:generalized-interpretation}
        \textbf{Comparing attack characteristics.}
        Existing works to interpret adversarial attacks on deep neural networks often focus on visualizing the activation patterns for a single adversarial input \cite{carter2019activation, norton2017adversarial}.
        We aim to visualize how the activation pathway changes as the attack strength varies, 
        to help users gain deeper insight into how the attack works generally.
        Understanding model vulnerability under different attack strategies informs more robust defenses \cite{das2018shield,papernot2016distillation,moosavi2016deepfool}.

    \item \label{challenge:access}
        \textbf{Lowering barrier of entry for interpreting and deciphering adversarial attacks.}
        The visualization community is contributing a variety of methods and tools to help people more easily interpret different kinds of DNNs \cite{bau2017network,liu2018analyzing,hohman2019summit,carter2019activation,olah2017feature,kahng2018gan,hohman2018visual}.
        Efforts that aim to support deciphering adversarial attacks, however, are relatively nascent
        \cite{bau2017network,liu2018analyzing, norton2017adversarial}.
        We aim to make interpreting adversarial attacks more accessible to everyone, following the footsteps of prior success from the community.
\end{enumerate}

\subsection{Background: Neuron Importance and Influence}
\label{sec:compute-neuron}

To discover activation pathways triggered by benign and adversarial inputs,
\model{} finds important neurons and influential connections among such neurons.
Inspired by \cite{hohman2019summit},
\model{} computes a neuron's \textit{importance} based on how strongly it 
is activated by all inputs, and the
\textit{influence} between neurons
based on the amount of activation signals transmitted through the connection to the next layer.
While summarizing interpretable pathways within a DNN 
remains an open problem
\cite{ilyas2019adversarial,brendel2019approximating,morcos2018importance},
recent works
\cite{olah2020zoom,hohman2019summit,olah2017feature}
have shown that dominant neuron activations at each layer
form the basis vectors for the entire activation space of the DNN.
Thus, characterizing \textit{important} neurons at each layer 
based on neuron activation
provides a surrogate sampling of important neurons 
across the whole network, for a given set of images.
\model{} extends this notion to scalably aggregate the activation pathways
across \textit{multiple} contexts 
with the most important neurons for: 
(1) benign images belonging to the \textbf{\green{original}} class, on which the targeted attacks are performed;
(2) benign images belonging to the \textbf{\blue{target}} class, which the attacks try to flip the label to; 
and (3) successfully \textbf{\red{attacked}} images for a particular attack strength (we support exploration with multiple strengths).

To begin, we consider the DNN model $\mathcal{M}$ (\inception{}),
where $Z^q \in \mathbb{R}^{H_q, W_q, D_q}$ is the output tensor 
of the $q$'th layer of $\mathcal{M}$.
Here, $H_q, W_q$ and $D_q$ are the 
height, width and depth dimensions respectively.
This implies that the layer has $D_q$ neurons. 
We denote the $d$'th output channel (for $d$'th neuron)
in the layer as $\mathcal{C}^d_q \in \mathbb{R}^{H_q, W_q}$.
We index the values in the channel
as $\mathcal{C}^d_q[h,w]$.
Given an input image $x_i$,
we find the maximum activation of each neuron induced by the image
using the global max-pooling operation:
$a^d_q[i] = \max_{h, w} \mathcal{C}^d_q[h, w]$. 
This represents the magnitude 
by which the $d$'th neuron in the $q$'th layer 
maximally detects the corresponding semantic feature from image $x_i$.
This technique of extracting maximal activation 
as a proxy for semantic features has also
demonstrated tremendous predictive power in the
data programming domain~\cite{das2020goggles}.
Finally, we pass all images from each of 
\textbf{\green{original}}, \textbf{\blue{target}} and \textbf{\red{attacked}} datasets.
For each set, we aggregate $a^d_q[i]$ values for all images 
and quantify the importance of each neuron
by the median value of such maximal activations.
We use medians for summarizing the neuron importance,
because they are less sensitive to extreme values.
Consistent with the findings of~\cite{hohman2019summit},
we observe that 
the maximal activation values are power law distributed,
implying that only a small minority of neurons have
highest importance scores.
Hence to denoise inconsequential visual elements, for each layer,
we empirically filter the 10 most important neurons for benign images of original and target classes,
and 5 most important ones for attacked images of each attack strength 
(i.e., at most 50 important neurons across all 10 attack strengths).

To compute influence of a connection between two neurons,
we measure the signal transmitted through the connection,
computed by the convolution of 
the slice of the kernel tensor between the two neurons 
over the source neuron's channel activation.
Since output from the ReLU activation function 
is used as the neuron activation in an \textsc{InceptionV1} model,
it implies that the neuron importance scores that are propagated are non-negative.
Consequently, the model acts on these non-negative activation values.
Hence, these activation values accumulate only for 
positively weighted convolution operations
through consecutive layers, which has the effect of filtering out 
non-influential connections 
that may even originate from important neurons.
Inspired by \cite{hohman2019summit}, 
we take the maximum value in the convolution
for the influence of the connection.
\model{} deviates from \cite{hohman2019summit}
when aggregating influence values across several images.
We characterize the connection
by taking the median influence across all images from a given set.
We take this approach since we want to summarize the influence characteristics
across multiple datasets (\textbf{\green{original}}, \textbf{\blue{target}} 
and \textbf{\red{attacked}} for different attack strengths), 
and each dataset is of a different size.
Simple counting may skew the results towards a particular dataset while 
the median value provides a characteristic aggregation of the influence scores.

\subsection{Realizing Design Goals in \model{}'s Interface}
\label{subsec:realize-bluff}

\model{}'s interface (\autoref{fig:teaser}) consists of:
\textbf{A. \sidebar{}} for selecting which data are included, filtered, highlighted, and compared;
\textbf{B. \graphview{}} that summarizes and visualizes \textit{activation pathways} as a graph;
\textbf{C. \detailview{}} for interpreting the concept that a neuron has detected, via feature visualization, representative dataset examples, and activation patterns over attack strengths.
In the header (\autoref{fig:teaser}, top), users can select a pair of  \textbf{\green{original}} and \textbf{\blue{target}} class.
\model{} then generates the main visualization in the \graphview{} for how neural networks misclassify \textbf{\green{original}} images as \textbf{\blue{target}} images when attacked, by displaying the activation pathways of adversarial inputs.

\smallskip
\noindent
\textbf{Unifying Multiple Graph Summaries.}
\model{} summarizes InceptionV1's responses to inputs under \textit{multiple} contexts in a unified view.
Specifically, the \graphview{} (\autoref{fig:teaser}B) visualizes 
the top neurons important only for the \textit{original} class as \textbf{\green{green}} nodes, 
those important only for the \textit{target} class as \textbf{\blue{blue}} nodes,
those important to both classes as \textbf{\orange{orange}} nodes,
and those important only for successfully-attacked images as \textbf{\red{red}} nodes.
Furthermore, those neurons are spatially grouped based on the four roles.
This generic design, that can be extended to any DNN model with intermediate convolutional blocks, helps users more easily pinpoint and tease out the subtle ways that neurons participate in an attack (\ref{challenge:entangled-pathways}, \ref{challenge:comparison}), e.g., red neurons, by the very definition, are exploited only by the attack but are neither activated by the original or target images.
Our key design decision here is to unambiguously differentiate the four neuron contexts using spatial positioning;
we supplement this differentiation by further encoding the four contexts with distinct colors.

The \graphview{} (\autoref{fig:teaser}B)
focuses on visualizing the model's 9 mixed layers (mixed3a, mixed3b ... mixed5b),
following existing interpretability literature~\cite{olah2017feature, olah2018building,hohman2019summit}.
The topmost row corresponds to the last mixed network layer (i.e., mixed5b).
Each connection between two neurons is visualized as a curved line, whose width scales linearly with the influence values computed as in \autoref{sec:compute-neuron}.

\smallskip
\noindent
\textbf{Visualizing exploited activation pathways.}
An adversarial input is often a slightly perturbed version of a benign input,
which means the activation pathways of an benign image and those of its adversarial counterpart would be  similar at the input layer ~\cite{liu2018analyzing},
yet decidedly different at the output layer --- the \textit{benign} pathways lead to 
the \green{\textbf{original}} prediction, 
while the \textit{adversarial} pathways lead to
the \blue{\textbf{target}} prediction.
Given the similar starting points but different outcomes, the adversarial activation pathways must have deviated from the benign pathways.
\model{} helps discover vulnerable neurons and connections that contribute to such deviations and the resulting misclassification,
by highlighting the neurons and connections that are \textit{excited} (or \textit{inhibited}, oppositely) the most by an attack (\autoref{fig:teaser}A) (\ref{challenge:entangled-pathways}).
A pathway \textit{excited} by an attack 
means its constituent neurons are activated more than expected (i.e., pathway contains more target features).
\autoref{fig:teaser} shows an example of where the attack \textit{excites} multiple 
features and connections to induce the target prediction of armadillo (e.g., ``\textit{scales},'' ``\textit{bumps},''  ``\textit{mesh},'' and ``\textit{brown bird}'' thanks to its similarity to armadillos' roundish, brown body).
Computationally, 
in layer $q$,
a neuron $d$'s excitation amount is 
$\tilde{a}^{q}_d[attacked] - \tilde{a}^{q}_d[benign]$, 
where $\tilde{a}^{q}_d[benign]$ and $\tilde{a}^{q}_d[attacked]$ 
are the neuron's importance for some benign and attacked images respectively (as described in \autoref{sec:compute-neuron}).

\smallskip
\noindent 
\textbf{Interpreting Activation Pathways.}
To help users more easily interpret the concepts that a neuron is detecting, alongside each neuron, \model{} shows 
(1) a \textit{feature visualization}, an algorithmically generated image that maximizes the neuron's activation, and 
(2) \textit{dataset examples}, cropped from real images in the dataset, 
that also highly activate the neuron \cite{olah2017feature}.
Hovering on a neuron shows the corresponding feature visualization and dataset examples as seen in \autoref{fig:teaser}C,
where adversarial images successfully 
induce the ``\red{\textit{brown bird}}'' feature, an appearance more likely shared by an armadillo (small, roundish, brown body) than a panda,
which in turn activates more features in subsequent layers that contribute to the (mis)classification of armadillo.
These visual explanations help translate abstract activation pathways into the composition and flow of learned concepts (\ref{challenge:comparison}).

\setlength{\intextsep}{0in}
\setlength{\columnsep}{0.1in}
\begin{wrapfigure}{R}{0.95in}
    \centering
    \includegraphics[width=0.95in]{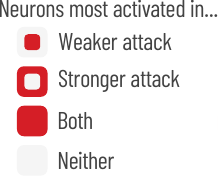}
    \vspace{-0.13in}
    \label{fig:compare-legend}
\end{wrapfigure}

\smallskip
\noindent 
\textbf{Comparing attacks with varying strengths.}
\model{} offers the \textit{Compare Attacks} mode that visualizes and compares 
the pathway differences between a weaker attack and 
a stronger attack (\ref{challenge:generalized-interpretation}). 
\model{} visually encodes the neurons based on which attack strengths they responded to, drawing inspiration from Alper et al \cite{alper2013weighted}.
Each neuron consists of an inner and an outer rectangle: 
the \textit{inner} rectangle is colored when the neuron is in the activation pathways of the \textit{weaker} attack; 
whereas the \textit{outer} rectangle is colored when the neuron is 
in the activation pathways of \textit{stronger} attack.
Thus, our design can visually encode all four possible comparison results in relative terms, enabling us to use hue to encode the four neuron contexts. In other words, the comparison mode’s visual encoding gracefully builds on and preserves Bluff’s overall visual design.
Our terminology for \textit{weaker} and \textit{stronger} attacks
are relative,
as we do not assert any explicit threshold
for weak or strong attack strengths.

\smallskip
\noindent
\textbf{Cross-platform deployment with standard web technologies.}
To support reproducible research and broaden its access, 
\model{} uses standard web technologies (HTML/CSS/JavsScript stacks, and D3.js)
and can be accessed from any modern 
web browser (\ref{challenge:access})
at \href{https://poloclub.github.io/bluff}{\texttt{\blue{https://poloclub.github.io/bluff}}}.
We ran all the backend code that computes neurons' importance and connections' influence 
on a NVIDIA DGX-1 workstation equipped with 8 GPUs (each with 32GB memory), 80 CPU cores, and 504GB RAM.

\section{Discovery Usage Scenarios}
\label{sec:scenario}
We now demonstrate how \model{} enhance 
the understanding of adversarial attacks 
and reveal attack strategies that confuse a DNN.
For our scenarios, we pick from the 1000 classes of the ImageNet dataset~\cite{ILSVRC15},
which consists of $\sim$1.2 million images.

\begin{figure}[t]
  \centering
  \includegraphics[width=0.55\linewidth]{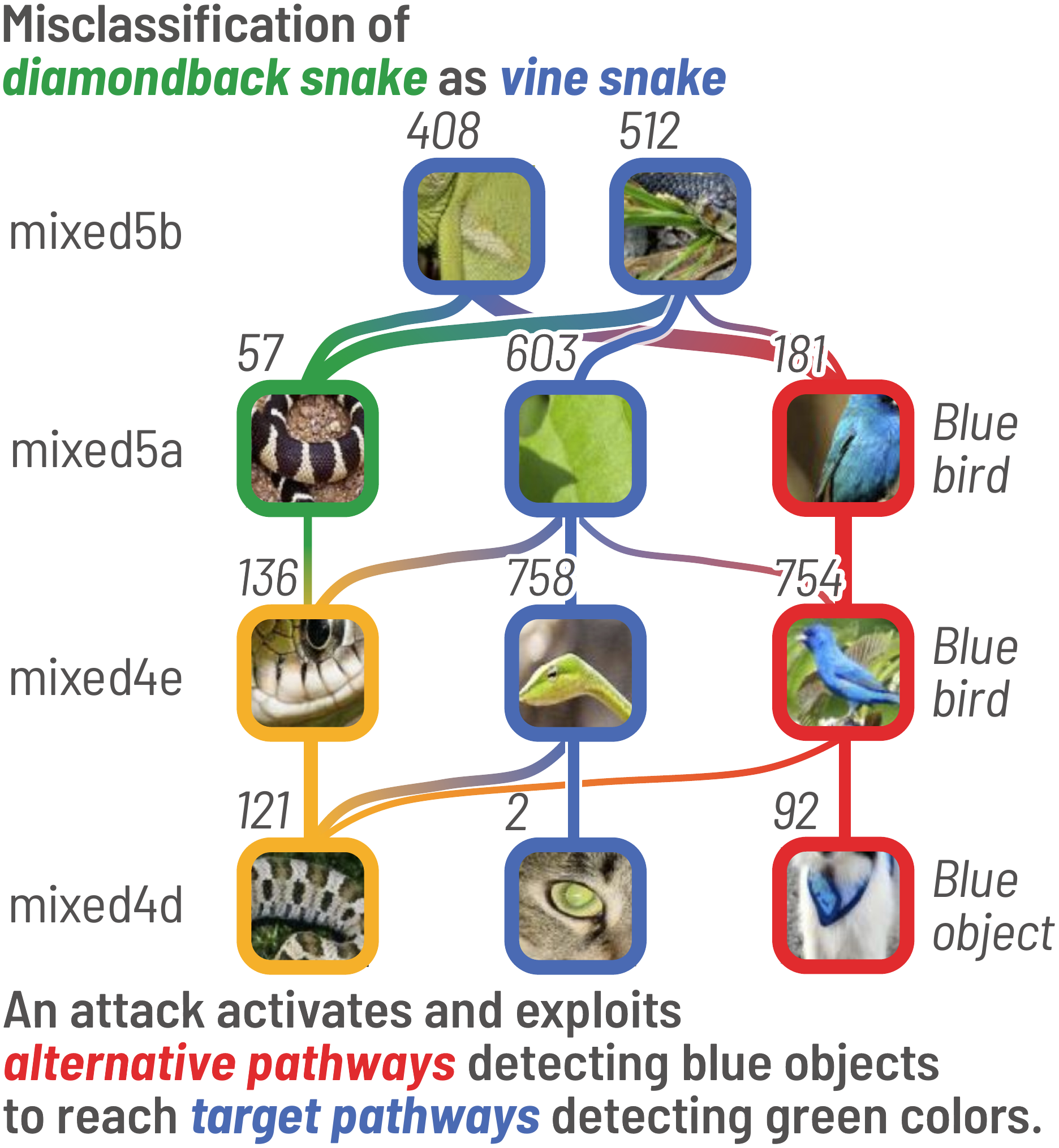}
  \vspace{-1em}
  \caption{
    \model{} helps users understand how an attack penetrates a model, by visualizing activation pathways that are additionally exploited by the attack.
    In this example, \model{} highlights the neurons and connections that PGD attack exploits (red) to make the model confuse adversarial \textbf{\green{diamondback snake}} images as \textbf{\blue{vine snake}}.
  }
  \label{fig:scenario-inside}
  \vspace{-1em}
\end{figure}

\subsection{Understanding How Attacks Penetrate DNNs}
\label{sec:penetrate}

Consider a DNN classifier that labels snakes,
such as the deadly venomous \textit{diamondback snake}, 
and the green \textit{vine snake} whose venom causes only mild swelling.
In \autoref{fig:scenario-inside}, \model{}'s \graphview{} reveals
how adversarial diamondback images \textit{exploit} (highly activate) 
unexpected pathways to induce the incorrect vine snakes prediction, leveraging multiple 
\red{exploited neurons} that look for \textit{``blue color''} (e.g., \textit{``blue birds"}
in \autoref{fig:scenario-inside}, right column).
This is surprising because \textit{vine snakes} (the attack's target class) have a green body, not blue.
This finding suggests that PGD exploits the pathway for \textit{``blue color''} 
as a bypassing alternative route to reach the pathways for vine snake,
which look for \textit{``green leaves''} and \textit{``green bumps''} 
(\autoref{fig:scenario-inside}, middle column).
We also noticed that PGD leverages \textit{``snake-like''} pathways 
that are important for \orange{both} classes (\autoref{fig:scenario-inside}, left column),
which is reasonable given that both the original and target classes are snakes.
Finding the pathways exploited by an attack provides 
fundamental insights that could inform future defenses,
such as blocking the alternative routes.
\autoref{fig:teaser} shows another example, where
the \textit{adversarial armadillo} pathways  overwhelm the \textit{benign panda} pathways and ultimately 
lead to the misclassification.

\subsection{Startling Tactic: Death by a Thousand Cuts}
\label{sec:strategy}

\begin{figure}[t]
  \centering
  \includegraphics[width=0.8\linewidth]{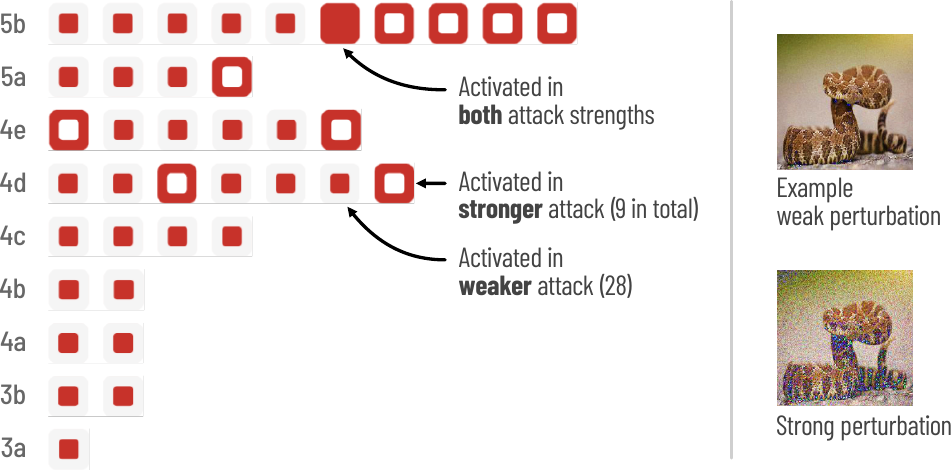}
  \vspace{-1em}
  \caption{
    Using \model{}'s “Compare Attacks” mode, 
    we examine PGD's different strategies
    for misclassifying \textit{diamondback} images into \textit{vine snake} class
    for two attack strengths (0.1 vs 0.5).
    The weaker attack exploits more alternative neurons (i.e., features that are not typically activated by benign inputs) than the stronger attack does.
    \vspace{-0.5em}
  }
  \label{fig:strong-weak-strategy}
  \vspace{-1em}
\end{figure}

\setlength{\intextsep}{0in}
\setlength{\columnsep}{0.1in}
\begin{wrapfigure}{R}{1.25in}
    \centering
    \includegraphics[width=1.25in]{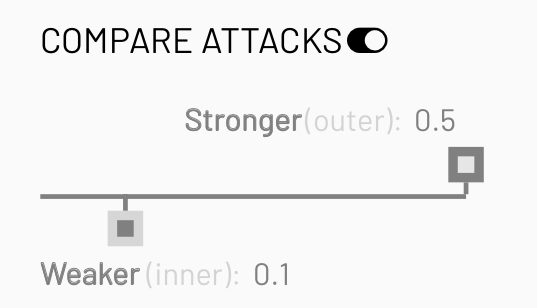}
    \vspace{-0.13in}
    \label{fig:compare-legend}
\end{wrapfigure}

An adversary can perform an attack on the model at various levels of attack strengths 
--- starting from imperceptible noise, all the way up to high intensity perturbation (see example perturbations in \autoref{fig:strong-weak-strategy}).
Does an attack's strategy evolves as the attack strength escalates,
or remains the same?
\model{} enables such a comparative analysis  through 
its \textit{Compare Attacks} mode.
Consider the example of attacking the \textit{diamondback} images 
to induce the misclassification of \textit{vine snake}.
Setting the weaker and stronger attack strength to 0.1 and 0.5 respectively
and looking at 30\% of most activated neurons in each layer,
\autoref{fig:strong-weak-strategy} reveals  
the surprising finding that the weaker attack 
exploits a large number of \red{red} neurons (28 in total) --- neurons that are \textit{not} important to either snake class, but are highly activated by the attack --- many more than the stronger attack (only 9 in total).
On further examining the example image patches for these neurons,
we noticed the images consist an assortment of semantic features
such as \textit{spider legs}, \textit{blue bird} and \textit{car hood},
seemingly unrelated to snakes.
We observed similar attack tactics in other class pairs.
For \textit{ambulance} images misclassified as \textit{street sign}, 
42 red neurons are exploited by the weaker attack, and only 16 by the stronger attack.
For \textit{panda} images misclassified as \textit{armadillo},
29 exploited by the weaker attack, only 8 by the stronger attack.
These observations lead us to conclude that weaker attacks
rely on leveraging a large number of disassociated semantic features to induce misclassification, i.e., ``death by a thousand cuts''.

\vspace{-0.3em}
\section{Discussion and Future Work}

We present \model{}, an interactive system for visualizing, characterizing, and deciphering adversarial attacks on DNNs.
We believe our visualization, summarization, and comparison approaches 
will help promote user understanding of adversarial attacks,
and support discoveries to design a proper defense.
Our next step is to use \model{} to help construct robust defenses against attacks.
We plan to extend \model{}
to support \textit{interactive neuron editing} (e.g., ``deleting'' a neuron from model),
so that the user may empirically identify and act on vulnerable neurons 
and observe the effects on the resulting pathway and prediction in real-time. 
We also plan to extend \textsc{Bluff} to work for adversarially-trained models
\cite{Goodfellow2014ExplainingAH, tramer2017ensemble, madry2018adversarial}, 
to help people gain deeper insights that explain their robustness.
Additionally, after receiving positive preliminary feedback from researchers, students and collaborators
who were given the opportunity to try out \textsc{Bluff},
we plan to  conduct user studies to evaluate our tool's usability and functionality.

\acknowledgments{
This work was supported in part by NSF grants IIS-1563816, CNS-1704701, NASA NSTRF, DARPA GARD, gifts from Intel, NVIDIA, Google, Amazon.
}

\bibliographystyle{abbrv-doi}

\bibliography{template}
\end{document}